\documentclass[11pt]{article}
\usepackage[utf8]{inputenc}
\DeclareUnicodeCharacter{2212}{-}

\usepackage[final]{acl}

\usepackage{booktabs}
\usepackage{siunitx}
\usepackage[table]{xcolor}

\usepackage{subcaption} 
\usepackage{titlesec}
\usepackage{times}
\usepackage{latexsym}
\usepackage{float}
\usepackage{placeins}

\usepackage{enumitem}
\usepackage{booktabs}
\usepackage{tabularx}
\usepackage{colortbl}
\usepackage{multirow}
\usepackage{siunitx}
\usepackage{caption}
\usepackage[T1]{fontenc}

\usepackage[utf8]{inputenc}

\usepackage{microtype}

\usepackage{inconsolata}

\usepackage{graphicx}

\title{Cross-Lingual Empirical Evaluation\\ of Large Language Models for Arabic Medical Tasks}

\author{
Chaimae Abouzahir,
Congbo Ma,
Nizar Habash,
Farah E. Shamout
\\[0.5em]
New York University Abu Dhabi \\
\texttt{\{ca2627,cm7196,nh48,fs999\}@nyu.edu}
}

\begin{document}
\maketitle
\begin{abstract}
In recent years, Large Language Models (LLMs) have become widely used in medical applications, such as clinical decision support, medical education, and medical question answering. Yet, these models are often English-centric, limiting their robustness and reliability for linguistically diverse communities. Recent work has highlighted discrepancies in performance in low-resource languages for various medical tasks, but the underlying causes remain poorly understood. In this study, we conduct a cross-lingual empirical analysis of LLM performance on Arabic \& English medical question and answering. Our findings reveal a persistent language-driven performance gap that intensifies with increasing task complexity. Tokenization analysis exposes structural fragmentation in Arabic medical text, while reliability analysis suggests that model-reported confidence and explanations exhibit limited correlation with correctness. Together, these findings underscore the need for language-aware design and evaluation strategies in LLMs for medical tasks.
\end{abstract}

\section{Introduction}

Large Language Models (LLMs) have shown remarkable performance on a wide range of medical tasks, including clinical question answering \citep{singhal2023}, medical reasoning \citep{chen2025, wu2025}, and exam-style benchmarks \citep{pal2022}, positioning them as as powerful capabilities for advancing healthcare applications. However, these successes are largely demonstrated in English due to limited availability of diverse benchmarks \citep{singh2025}.

As LLMs move closer to real-world healthcare deployment, their ability to function reliably across languages becomes a critical concern. Recent multilingual evaluations consistently report substantial performance drops when medical LLMs are evaluated outside English, with Arabic as one of the affected languages \citep{Alonso_2024}. However, reported results are typically limited to aggregate performance scores \citep{medarabiq}, providing limited insight into the underlying causes of model underperformance.

Despite growing recognition of performance gaps between Arabic and English, existing explanations remain largely underexplored, often attributing failures to limited pretraining data or domain mismatch \citep{jin2023, qiu2024multi}. As a result, it remains unclear whether poor Arabic performance stems primarily from linguistic properties of the language, insufficient medical domain adaptation, architectural design choices, or interactions between these factors. This hinders principled adaptation: \textbf{without knowing which factors dominate model failure, it is difficult to design effective multilingual training strategies, alignment procedures, or evaluation protocols.}
To this end, we present the first systematic study designed to disentangle linguistic, domain-specific, and architectural contributors to LLM performance on Arabic medical tasks. We make the following contributions:
\begin{itemize} [topsep=2pt, itemsep=2pt]
    \item We design a cross-lingual diagnostic evaluation framework for general-purpose and medical LLMs that enables controlled analysis across languages, output formats, tokenization behavior, and reliability signals.
    
    \item We conduct an empirical study on MedAraBench, an Arabic medical question answering dataset, and its English-translated counterpart to isolate language effects while controlling for medical content.
    \item Our findings show that Arabic performance degradation is driven by interacting representational, alignment, and evaluation factors rather than medical knowledge alone, with gaps amplifying under increased task complexity and free-form generation.
\end{itemize}


\section{Related Work}
\subsection{Medical LLMs and Evaluation Benchmarks}

LLMs have driven recent progress in clinical NLP, supporting applications including decision support, diagnostic assistance, and clinical text generation. To assess medical reasoning capabilities, several evaluation benchmarks, primarily formulated as question-answering tasks based on medical examinations or curated clinical sources, have been introduced \citep{zhang2018, jin2019, pal2022}. Benchmarks such as MedQA and PubMedQA \citep{jin2020,jin2019} are now widely used to evaluate medical knowledge and reasoning in LLMs.

General-purpose LLMs have achieved strong performance on English medical benchmarks. Notably, GPT-4 exceeded the passing threshold on USMLE-style questions in MedQA, achieving an accuracy of 86.1\% \citep{nori2023}. This success motivated the development of medical-domain LLMs through domain-specific adaptation. Proprietary models such as Med-PaLM and Med-PaLM 2 \citep{singhal2023}, as well as GPT-4 MedPrompt \citep{nori2023generalist}, reported substantial gains, with GPT-4 MedPrompt surpassing 90\% accuracy on MedQA and achieving significant error reduction.

However, the costs, opacity, and privacy constraints associated with proprietary systems have limited their adoption in real-world clinical settings. In response, several open-access medical LLMs have been proposed, yet their performance on established benchmarks remains limited. Unlike proprietary models, BioMistral only achieves 44.4\% accuracy on MedQA, while MedAlpaca and PMC-LLaMA attain 35.4\% and 27.6\%, respectively \citep{labrak2024-biomistral}. These results highlight a persistent performance gap between proprietary and open-source medical LLMs.


\subsection{Multilingual Medical Benchmarks and Cross-lingual Generalization}

Despite the widespread evaluation of LLMs on English medical benchmarks, their reliability across languages remains limited. Prior work has shown that both general-purpose and medical LLMs are prone to hallucinations \citep{xiong2024hallucinate} and may produce answers based on outdated clinical knowledge \citep{vladika2025outdated}. Moreover, most medical benchmarks are predominantly English-centric in both their construction and evaluation \citep{qiu2024multi}.



Recent multilingual evaluations consistently report substantial performance drops outside English. For instance, significant degradation has been observed on Italian medical QA tasks  \citep{kembu2025italian}, as well as across a broader range of non-English healthcare queries \citep{jin2023}. \citet{Alonso_2024} further show that both general-purpose and medical LLMs perform markedly worse in Arabic and Hindi than in English. Notably, medical LLMs often underperform the base models from which they are adapted in non-English settings, suggesting that domain adaptation may reduce cross-lingual generalization.

\citet{Alonso_2024} further show that medical LLMs often underperform their base models in non-English settings such as Arabic and Hindi, suggesting that domain adaptation may hinder cross-lingual generalization. Complementarily, \citet{jeong2025} demonstrate that this effect already occurs in English, indicating that specialization alone does not guarantee performance gains even without language mismatch.

To mitigate these disparities, some efforts have focused on developing language-specific medical models. HuatuoGPT \citep{zhang2023} is a notable example of a Chinese medical LLM trained on native-language biomedical resources. However, systematic analyses of how domain adaptation interacts with multilingual performance remain limited, particularly for underrepresented languages. 

\subsection{Challenges in Arabic Medical Language Models}

Arabic poses distinct challenges for medical language modeling, including rich morphology, complex tokenization, dialectal variation, and a scarcity of high-quality, domain-specific resources \cite{farghaly2009,Habash:2010:introduction}. Although general-purpose Arabic LLMs such as Jais \citep{sengupta2023jais}, Fanar \citep{fanarteam2025} and ALLAM \citep{bari2024} have been introduced, the development and evaluation of Arabic medical LLMs remain underexplored.

Existing evaluations report poor performance on Arabic medical tasks \citep{medarabiq}. However, the underlying causes of these failures are not well understood. It remains unclear whether performance degradation primarily arises from linguistic representation issues, limitations of domain-adaptive training, or their interaction, particularly for medical LLMs adapted from English-centric base models. Moreover, the lack of publicly available Arabic medical benchmarks limits systematic diagnostic analyses comparable to those in English \citep{alasmari2024}, motivating our investigation into the mechanisms underlying Arabic medical LLM failures beyond aggregate performance.

\section{Methodology} 

To investigate sources of performance degradation in Arabic medical MCQs, we design a targeted evaluation framework probing LLM behavior on several aspects. Rather than introducing a new model, we focus on a set of research questions, which we detail below.

\subsection{Research Questions}

We compare model accuracy on original Arabic questions and their English-translated counterparts to isolate the role of linguistic representation from medical reasoning.

\noindent \textbf{RQ1: To what extent is performance degradation driven by language rather than medical reasoning?}  
We compare model accuracy on original Arabic questions and their English-translated counterparts to isolate the role of linguistic representation from medical reasoning.

\noindent  \textbf{RQ2: How do question-level properties affect model performance?}  
We analyze accuracy as a function of input length, question difficulty, and medical specialty to determine whether linguistic complexity, cognitive demand, or domain-specific content disproportionately affects model outcomes.

\noindent \textbf{RQ3: How do alignment constraints and output formats influence model behavior across languages?}  
We compare soft matching (letter-based option selection) and hard matching (exact answer text generation) to evaluate how instruction following and surface-form generation influence accuracy across languages.

\noindent \textbf{RQ4: Does tokenization behavior contribute to Arabic performance gaps?}  
We examine tokenizer efficiency and fragmentation patterns to understand whether Arabic morphology and segmentation lead to less effective input representations.

\noindent  \textbf{RQ5: Are model confidence estimates and generated explanations reliable indicators of correctness?}  
We analyze model-reported confidence and accompanying rationales to assess whether they correlate with accuracy and can be used to diagnose systematic failure modes.

\subsection{Dataset}
All experiments are conducted on MedAraBench \cite{abudaoud2026medarabenchlargescalearabicmedical}, an Arabic medical question answering benchmark. The questions are originally authored in Modern Standard Arabic (MSA), collected from medical exams, digitized from scanned paper sources, and manually curated to exclude incomplete or ambiguous items. Each question is annotated with the number of answer options (4–6), a difficulty level corresponding to years of medical study (Y1–Y5), and a medical specialty. The dataset covers 19 specialties (e.g., Anatomy, Pathology, Surgery, Pharmacology) and is split into training and test sets using an 80/20 split with matched specialty distributions (19,894 train / 4,989 test). A data sample is shown in Appendix \ref{fig:data-sample}. 

Models are evaluated on a medical MCQ task in both Arabic and English. English versions are obtained via automatic translation of the original Arabic questions using the Google Translate API and are used solely for controlled cross-lingual analysis. Models are evaluated using accuracy, with a prediction counted as correct if the selected option matches the gold label.

\subsection{Evaluated Models}

We evaluate several recent open-source large language models as baselines for Arabic medical MCQs. We include recent, large-scale general-purpose LLMs like DeepSeek-V3.2 and LLaMA 3.3 70B as representative contemporary baselines in our evaluation. To examine differences between general language ability and domain-specific modeling, we compare two categories of models:

\begin{itemize}
\item \textbf{General-purpose LLMs}: DeepSeek-V3.2 \citep{deepseekv32}, LLaMA 3.3 70B \citep{llama33_70b}, and Mistral-Small-3.2-24B-Instruct-2506 \citep{mistral_small32_24b}, all trained on broad multilingual or mixed-domain corpora.
\item \textbf{Medical-domain LLMs}: Meditron 3 70B \citep{meditron3_70b_2024}, Med42-70B \citep{med42}, and MedGemma-27B-text-it \citep{sellergren2025}, which incorporate domain-adaptive pretraining or finetuning on medical data. None explicitly report multilingual medical pretraining, and available documentation indicates predominantly English medical data.
\end{itemize}


Moreover, our evaluation focuses exclusively on open-source models for both methodological and practical reasons. From a methodological standpoint, open-source models offer full access to embeddings, tokenizers, and intermediate representations, which are essential for both our analysis and our planned cross-lingual adaptation method. For practical concerns, deploying black-box proprietary systems in medical settings poses significant privacy and auditability concerns, underscoring the need for transparent, open-source alternatives.

\subsection{Experimental Settings}

All models were evaluated using a unified multiple-choice prompting setup implemented via the HuggingFace Transformers API. Inference used greedy decoding (temperature = 0, no sampling, top-p = 1.0, top-k disabled). We fix task-specific maximum generation lengths across languages to ensure comparable inference conditions across models, allowing up to 4 tokens for letter matching, 15 for text matching, and 70 for explanation generation. These limits were chosen to accommodate the required output formats rather than language-specific tokenization characteristics. 

The system prompts used follow standardized MCQ templates as shown in Appendix \ref{app:prompts} and are provided in English for all inputs, including Arabic inputs. This is based on preliminary prompt-engineering experiments showing more stable and higher-performing outputs than Arabic prompts. This choice reflects the English-centric instruction-following capabilities of the evaluated models and results in mixed-language inputs for Arabic evaluations.

Due to hardware constraints, all 70B-parameter models were evaluated using 4-bit NF4 quantization with bfloat16 compute (BitsAndBytes) across two 32 GB V100 GPUs. Smaller models, including Medgemma-27B-text-it and Mistral-Small-3.2-24B-Instruct-2506, were evaluated without quantization in full bfloat16 precision on the same hardware. DeepSeek-V3.2 could not be evaluated locally due to its size and was instead accessed via the official DeepSeek API. For cross-lingual analysis, models were additionally evaluated on an English-translated version of the dataset. This setup enables direct comparison between Arabic and English under identical task structures, isolating the effect of language from content.

\section{Empirical Studies and Analyses}

\begin{table}[!t]
    \centering
    \small
    \sisetup{mode=text}
    \setlength{\tabcolsep}{3pt}
    \renewcommand{\arraystretch}{0.95}
    \begin{tabular}{l c c c}

        \toprule
        \textbf{Models} & {\textbf{Acc (Ar)}} & {\textbf{Acc (En)}} & {\textbf{$\Delta$ (En–Ar)}} \\
        \midrule
        \rowcolor{gray!12}
        \multicolumn{4}{@{}c@{}}{\emph{General-purpose LLMs}} \\
        DeepSeek-V3.2 &  \textbf{\num{62.39}}  & \textbf{\num{62.85}} &  \textbf{\num{0.46}}\\
        Llama 3.3 70B & 42.10 & 57.61 & 15.51\\
        Mistral-Small-3.2-24B & 50.25 & 57.75 & 7.5\\
        \midrule
        \rowcolor{gray!12}
        \multicolumn{4}{@{}c@{}}{\emph{Medical-domain LLMs}} \\
        Meditron 3 70B & 50.51 & 58.80 & 8.92\\
        Med42-70B & 33.59 & 53.21 & 19.62\\
        medgemma-27b-text-it & 49.22 & 52.30 & 3.08\\
        \bottomrule
    \end{tabular}
    \caption{\textbf{Results of general-purpose and medical-domain LLMs' Acc(uracy) on the Arabic (Ar) and English (En) datasets.} Bold values indicate the highest accuracy within each column. Mistral-Small-3.2-24B refers to Mistral-Small-3.2-24B-Instruct-250.}
    \label{tab:main}
\end{table}

\subsection{Assessing the Role of Language in LLM Performance}

We compare model accuracy on parallel English and Arabic medical benchmarks using a controlled prompting setup (Appendix \ref{fig:prompt-letter}) to isolate the effect of linguistic representation from medical reasoning. As shown in Table \ref{tab:main}, accuracy is consistently lower in Arabic than in English across nearly all evaluated models, indicating a systematic language-associated performance gap. DeepSeek-V3.2 is the only model that exhibits comparable performance across languages. Notably, this behavior is not observed uniformly in larger models, indicating that reduced cross-lingual degradation cannot be attributed to model size alone.

For models with comparable parameter sizes ($\leq 70\text{B}$), English consistently outperforms Arabic, indicating that language remains a key factor even under similar capacity constraints. This trend holds for both general-purpose and medical-domain models, suggesting that domain specialization alone does not resolve Arabic performance gaps.

\begin{figure*}[t]
  \centering
  \includegraphics[width=\textwidth]{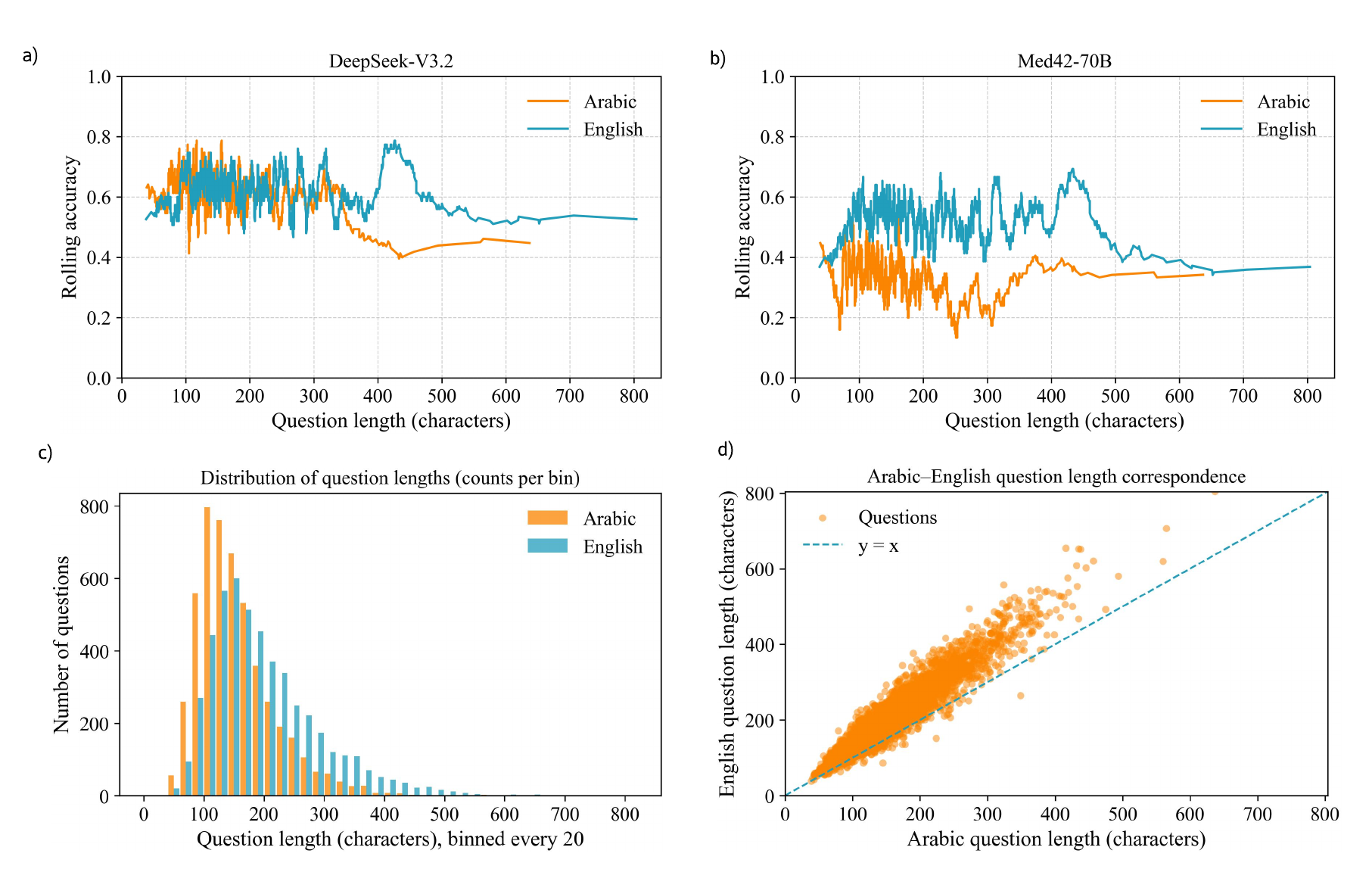}
  \caption{\textbf{Effect of question length on accuracy across Arabic and English.} (a–b) Rolling accuracy versus question length for DeepSeek-V3.2 and Med42-70B, respectively. (c) Distribution of question lengths in both languages. (d) Arabic–English length correspondence for aligned question pairs.}
  \label{fig:length}
  \vspace{10pt}
\end{figure*}

\subsection{Effects of Question Length and Difficulty}

We investigate whether question-level characteristics influence model accuracy by analyzing performance trends with respect to input length, educational difficulty, and medical specialty. We focus our analysis on the best- and worst-performing models overall, DeepSeek-V3.2 and Med42-70B, respectively, and restrict the following experiments to these two models.

Figures~\ref{fig:length} (a,b) show accuracy trends as a function of question length for DeepSeek-V3.2 and Med42-70B. Accuracy is relatively stable for shorter inputs but degrades as question length increases for Arabic, while English performance remains comparatively stable at longer lengths. Question lengths are strongly correlated across paired Arabic–English items and exhibit overlapping distributions (Figures~\ref{fig:length} c,d), indicating that the observed degradation reflects increased sensitivity to input length rather than artifacts of translation or systematic length mismatches.

\begin{figure}[!t]
    \centering
    \includegraphics[width=\linewidth]{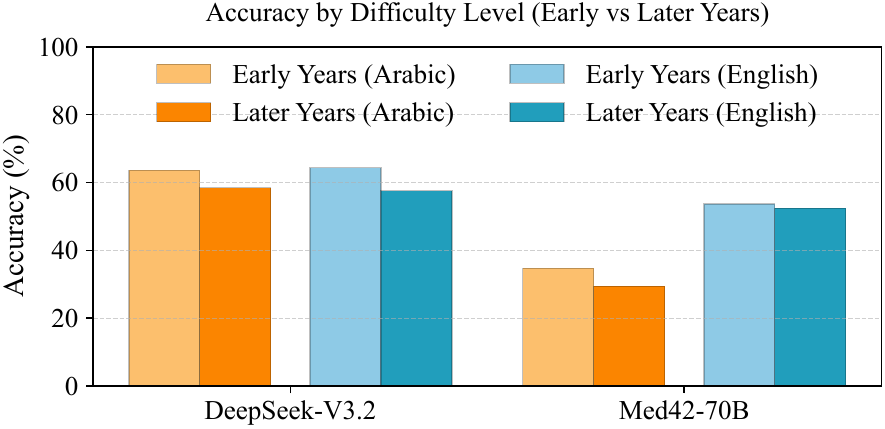}
    \caption{\textbf{Accuracy by educational difficulty level (early vs. later years) for DeepSeek-V3.2 and Med42-70B on Arabic and English medical MCQs.}}
    \label{fig:accuracy_by_diff}
\end{figure}

Figure \ref{fig:accuracy_by_diff} reports accuracy by educational difficulty level. For both models and languages, accuracy decreases for later years' questions (Y3+) compared to early years' questions (Y1–Y2). The performance drop is consistently larger for Arabic, particularly for Med42-70B.

\begin{figure}[!t]
    \centering
    \includegraphics[width=\linewidth]{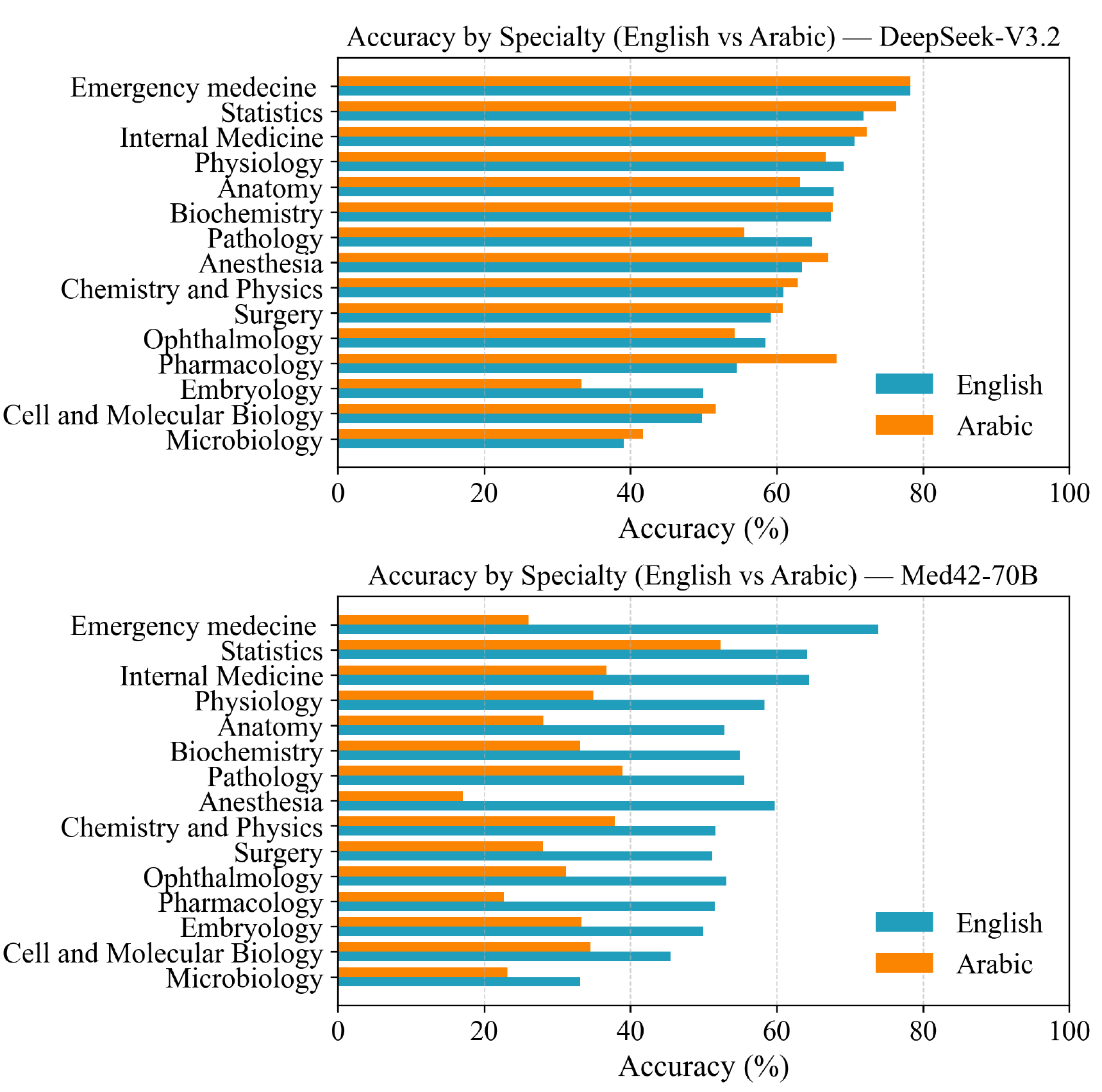}
    \caption{\textbf{Accuracy by medical specialty for DeepSeek-V3.2 (top) and Med42-70B (bottom) on Arabic and English medical MCQs.}}
    \label{fig:accuracy_by_specialty}
    \vspace{20pt}
\end{figure}

Figure~\ref{fig:accuracy_by_specialty} shows accuracy by medical specialty, revealing substantial variation across domains: performance is higher in clinically oriented fields (e.g., Emergency Medicine, Internal Medicine) and lower in foundational or detail-intensive specialties such as Microbiology and Embryology. English consistently outperforms Arabic across most specialties. This gap is particularly pronounced for Med42-70B, where Arabic performance lags behind English across nearly all specialties, suggesting that language-related performance disparities persist even when controlling for domain. 

Overall, these results indicate that input length, difficulty, and domain content systematically affect model performance, and that these effects disproportionately impact Arabic compared to English.

\subsection{Alignment Behavior Analysis}

\begin{figure}[!t]
    \centering
    \includegraphics[width=\linewidth]{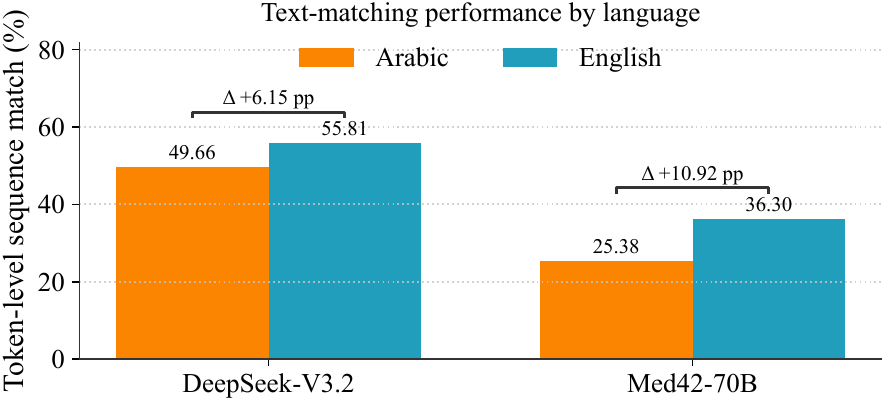}
    \caption{\textbf{Token-level sequence-match accuracy (\%) for text-matching evaluation in Arabic and English across two models.}}
    \label{fig:alignment}
\end{figure}

To analyze how output format influences model behavior across languages, we evaluate model performance under free-form answer generation using the prompt in Appendix \ref{fig:prompt-text}. We use token-level sequence similarity between predicted and ground-truth answer texts computed with the SequenceMatcher algorithm \citep{sequencematcher}. Figure \ref{fig:alignment} shows that across both models, surface-form similarity is consistently lower for Arabic than for English, with a larger gap of 10.92 percentage points for Med42-70B compared to 6.15 pp for DeepSeek-V3.2. These discrepancies indicate that, when models are required to generate answer text explicitly, Arabic outputs diverge more substantially from reference answers at the surface-form level. The magnitude of these gaps is also larger than that observed under letter-based option selection reported in Table \ref{tab:main}, suggesting that free-form generation amplifies language-specific difficulties beyond those captured by standard MCQ accuracy.

\begin{table}[!t]
\centering
\small

\begin{subtable}{\columnwidth}
\centering
\setlength{\tabcolsep}{2pt}
\renewcommand{\arraystretch}{1.05}

    \begin{tabular}{l c c c}
\toprule
\textbf{Tokenizer} & {\textbf{Tok/Word}} & {\textbf{Char/Tok}} & {\textbf{Single Char}} \\
\midrule
\rowcolor{gray!12}\multicolumn{4}{@{}c@{}}{\emph{Model-native tokenizers}} \\

DeepSeek-V3.2 & {2.39} & {2.33} & {32\%} \\
Llama 3.3 70B & {2.42} & {2.27} & {32\%} \\
Meditron 3 70B & {2.45} & {2.27} & {32\%} \\
Mistral-Small-3.2-24B & {2.06} & {2.72} & {0\%} \\
Med42-70B & {2.42} & {2.27} & {32\%} \\
\midrule
\rowcolor{gray!12}\multicolumn{4}{@{}c@{}}{\emph{Multilingual-efficient tokenizer}} \\
Gemma-3-4B-it & {2.30} & {2.43} & {35\%} \\
\midrule
\rowcolor{gray!12}\multicolumn{4}{@{}c@{}}{\emph{Arabic-focused tokenizer}} \\
CAMeLBERT-MSA & {1.76} & {3.21} & {36\%} \\
\bottomrule
\end{tabular}
\caption{Arabic dataset.}
\label{tab:tok-ar}
\end{subtable}

\vspace{0.6em} 

\begin{subtable}{\columnwidth}
\centering
\setlength{\tabcolsep}{2pt}
\renewcommand{\arraystretch}{1.05}

 \begin{tabular}{l c c c}
\toprule
\textbf{Tokenizer} & {\textbf{Tok/Word}} & {\textbf{Char/Tok}} & {\textbf{Single Char}} \\
\midrule
\rowcolor{gray!12}\multicolumn{4}{@{}c@{}}{\emph{Model-native tokenizers}} \\

DeepSeek-V3.2 & {1.52} & {4.01} & {28\%} \\
Llama 3.3 70B & {1.60} & {3.82} & {27\%} \\
Meditron 3 70B & {1.60} & {3.82} & {27\%} \\
Mistral-Small-3.2-24B & {1.56} & {3.93} & {0\%} \\
Med42-70B & {1.60} & {3.82} & {27\%} \\
\midrule
\rowcolor{gray!12}\multicolumn{4}{@{}c@{}}{\emph{Multilingual-efficient tokenizer}} \\
Gemma-3-4B-it & {1.57} & {3.90} & {35\%} \\
\midrule
\rowcolor{gray!12}\multicolumn{4}{@{}c@{}}{\emph{Arabic-focused tokenizer}} \\
CAMeLBERT-MSA & {2.82} & {2.13} & {44\%} \\
\bottomrule
\end{tabular}
\caption{English dataset.}
\label{tab:tok-en}
\end{subtable}

\caption{\textbf{Tokenization fragmentation statistics for Arabic and English inputs}. We report average tokens per word (subword splitting), average characters per token (compactness), and single-character tokens (reflecting extreme fragmentation).}
\label{tab:tokenization-analysis}
\end{table}

\subsection{Tokenization Efficiency Analysis}

We analyze tokenization efficiency and fragmentation to assess whether language-specific tokenization patterns are associated with downstream performance gaps. We report average tokens per word, characters per token, and the proportion of single-character tokens in Table \ref{tab:tokenization-analysis}. For Arabic, word counts are computed using a linguistically informed tokenizer from CAMeL Tools~\cite{Obeid:2020:camel}, while English uses whitespace-based word segmentation. Higher tokens per word and single-character rates, together with lower characters per token, indicate more fragmented representations. Higher tokens per word and single-character rates, together with lower characters per token, indicate more fragmented representations.

Across model-native tokenizers, Arabic is consistently more fragmented than English, with approximately 2.4 tokens per word compared to 1.5–1.6 for English. Similar trends hold for the multilingual tokenizer. This increased fragmentation leads to higher token usage for Arabic inputs, which may plausibly contribute to sharper performance degradation as input length and complexity increase. However, this effect is not uniform across models: despite higher token counts for Arabic, some models (e.g., DeepSeek-V3.2) exhibit more stable performance, suggesting that tokenization alone does not fully explain the observed degradation. In contrast, the Arabic-focused CAMeLBERT tokenizer \cite{inoue-etal-2021-interplay} substantially reduces fragmentation for Arabic while increasing fragmentation for English, illustrating that tokenizer efficiency is language-dependent.

\begin{figure}[!t]
    \centering
    \includegraphics[width=\linewidth]{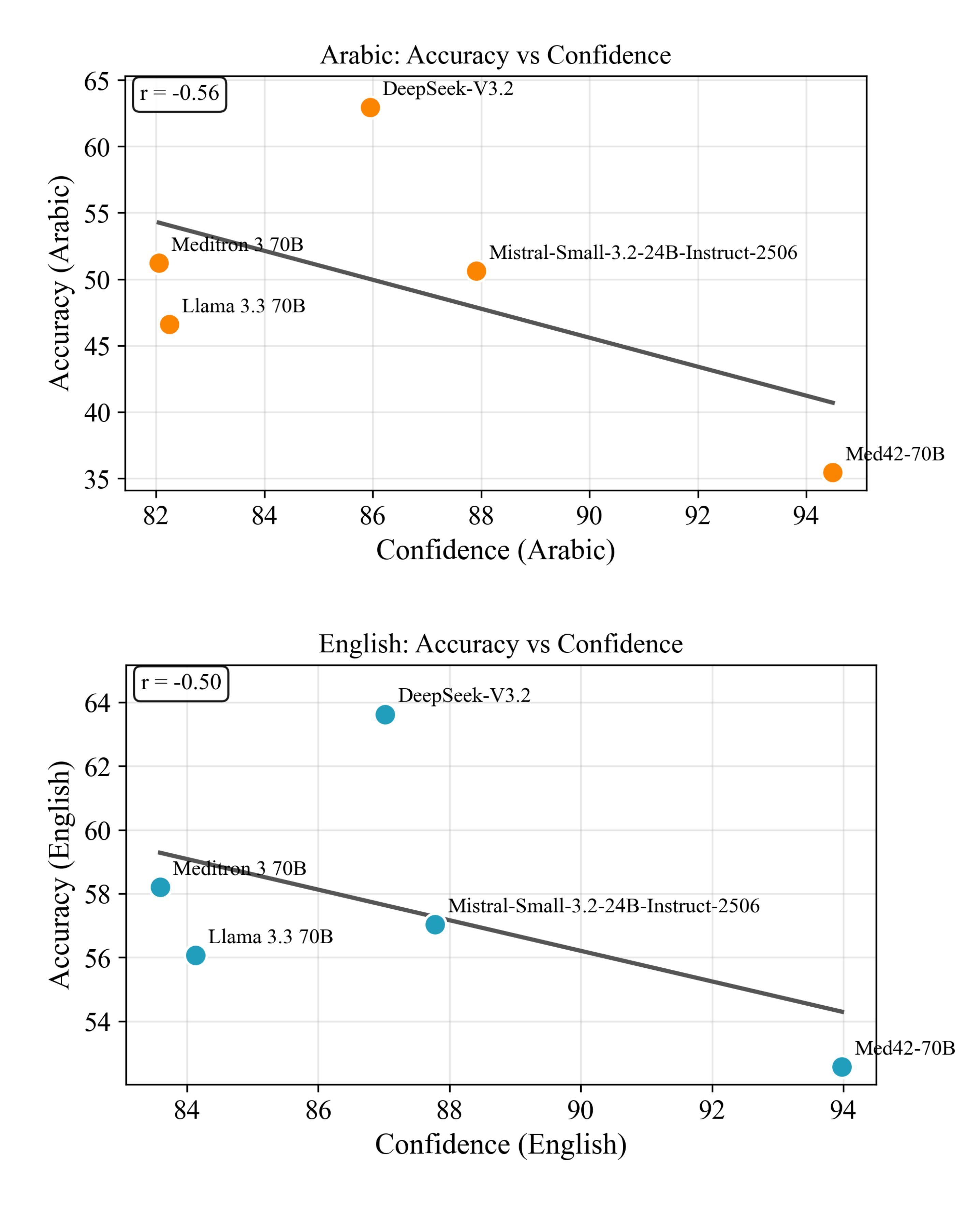}
    \caption{\textbf{Relationship between model-reported confidence and accuracy for Arabic (top) and English (bottom) multiple-choice medical MCQ.}}
    \label{fig:accuracy_confidence}
\end{figure}

\subsection{The Role of Confidence Estimates and Explanations}

We analyze the relationship between model-reported confidence (prompt in Appendix \ref{fig:prompt-conf}) and accuracy in medical question answering (Figure~\ref{fig:accuracy_confidence}). We observe a moderate negative correlation between confidence and accuracy in both Arabic ($r=-0.56$) and English ($r=-0.50$), indicating that higher confidence predictions are, on average, less accurate. This pattern is consistent across model families and languages, suggesting a general miscalibration of confidence in medical settings. Accordingly, model-reported confidence should not be treated as a reliable proxy for correctness. MedGemma-27B-text-it is excluded due to repeated noncompliance with the required confidence-reporting format under zero-shot prompting, often producing free-form text without a valid answer label. As this reflects instruction-following issues rather than task performance, we omit it from this analysis.

To examine whether explicit reasoning improves performance, we prompt models to generate a natural language explanation before answer selection, following a chain-of-thought–style prompting strategy \citep{wei2023} using the prompt shown in Appendix~\ref{fig:prompt-explanation}. MedGemma-27B-text-it is excluded for severe instruction non-compliance, consistent with the confidence-based analysis. As shown in Table \ref{tab:explanation}, explanation prompting yields mixed and often detrimental effects across models. While some models exhibit modest gains in Arabic accuracy (e.g., DeepSeek-V3.2: 62.39 → 63.56), explanation prompting often leads to degraded performance in English and, for several models, substantial drops in both languages (e.g., Meditron-3-70B and Med42-70B). 

Upon qualitative inspection, we find that in many cases, the model produces medically plausible or partially correct explanations while selecting an incorrect option label. Representative examples illustrating these reasoning–label misalignments are provided in Appendix \ref{fig:expl-samples}. Requiring explanations appears to encourage verbose reasoning and reinterpretation, which can decouple reasoning quality from discrete multiple-choice selection.
While explanation-conditioned prompting reduces accuracy for some models, this setting also surfaces cases where letter-only evaluation may overestimate performance by rewarding option matching despite stem–option inconsistencies. In such cases, explanation prompting exposes reasoning–label misalignment rather than pure knowledge errors.

\begin{table}[t!]
\centering
\small
\setlength{\tabcolsep}{2.5pt}
\begin{tabular}{lcccc}
\toprule
\textbf{Model} &
\textbf{Base$^{Ar}$} &
\textbf{Base$^{En}$} &
\textbf{Exp$^{Ar}$} &
\textbf{Exp$^{En}$} \\
\midrule
DeepSeek-V3.2  &  \textbf{62.39} & \textbf{62.85} & \textbf{63.56} & 45.85 \\
Llama-3.3-70B   & 42.10 & 57.61 & 46.27 & 57.91 \\
Mistral-Small-3.2-24B & 50.25 & 57.75 & 49.50 & \textbf{57.99} \\
Meditron-3-70B & 51.05 & 58.45 & 28.65 & 33.65\\
Med42-70B   & 33.59 & 53.21 & 28.99 &  33.41\\
\bottomrule
\end{tabular}
\caption{\textbf{Accuracy with Explanation prompting (Exp) compared to the baseline (Base, no explanation prompt), for Arabic and English}. Bold values denote the highest accuracy for each column.}
\label{tab:explanation}
\end{table}

\section{Discussion}
\subsection{Language as a Source of Degradation}

Across nearly all evaluated models, performance on Arabic medical MCQs is consistently lower than on their English-translated counterparts, despite identical medical content, indicating a persistent language-related performance gap beyond medical knowledge alone. DeepSeek-V3.2 is a notable exception, achieving near-parity between Arabic and English, demonstrating that strong cross-lingual performance in Arabic medical QA is achievable in open models.

This robustness cannot be explained by model scale alone, as public estimates suggest that DeepSeek-V3.2 and LLaMA 3.3 70B are trained at comparable orders of magnitude in compute and training tokens \citep{epochai}, yet only DeepSeek-V3.2 maintains high Arabic performance. This suggests that language robustness depends on specific design and training choices beyond scale, including data curation, language balance, and post-training procedures.

\subsection{Interaction between Task Complexity and Language}

Performance across all evaluated models is systematically higher for shorter questions and for lower-difficulty (early-year) items, regardless of language. However, the rate of performance degradation differs substantially between Arabic and English. As question length increases and as questions progress from early to later years' material, accuracy declines more sharply for Arabic than for English, as shown in Figures \ref{fig:length} (a-b) and \ref{fig:accuracy_by_diff}.

The relatively strong performance on short and early-year Arabic questions indicates that models can successfully answer simpler medical queries in Arabic, meaning that basic medical knowledge is present. The decline observed for longer and more advanced questions points to a reduced robustness of Arabic representations as task complexity increases, rather than a lack of medical understanding. A similar pattern emerges across medical specialties in Figure \ref{fig:accuracy_by_specialty}, where the gap between Arabic and English is larger in specialties that involve finer-grained distinctions, 
reinforcing 
the interaction between language effects and task complexity.

\subsection{Alignment Constraints and Evaluation Sensitivity}

The alignment analysis shows that output format influences how language-specific performance differences manifest. Because letter-based MCQ accuracy and token-level text similarity measure distinct aspects of model behavior, we restrict our analysis to within-format comparisons between Arabic and English. Under free-form answer generation, models consistently achieve lower token-level similarity in Arabic than in English, yielding larger cross-lingual gaps than those observed with constrained option selection. This suggests that removing output constraints introduces additional language-dependent variability not reflected by letter-based evaluation. In particular, free-form generation places greater demands on lexical choice and morphological realization, which are more challenging in Arabic. Since token-level similarity measures surface-form overlap, lower scores primarily reflect increased variation in answer expression rather than incorrect medical reasoning.

\subsection{Tokenization as a Structural Constraint} 

Compared to English, Arabic medical text is consistently more fragmented under model-native tokenizers, with words split into more subword units and a higher prevalence of single-character tokens. This reflects a mismatch between subword tokenizers, optimized for frequent training forms, and Arabic medical vocabulary, which combines rich morphology with low-frequency, variable domain-specific terms, leading to unstable subword representations and finer-grained splits.

Multilingual tokenizers partially mitigate this effect by covering broader lexical distributions, while Arabic-focused tokenizers further reduce fragmentation by explicitly modeling Arabic morphology. This shows how tokenizer training implicitly prioritizes certain linguistic distributions in ways that disadvantage underrepresented languages. Such fragmentation imposes a structural constraint on downstream processing: longer effective input sequences reduce usable context and increase sensitivity to question length, offering a plausible explanation for the sharper performance degradation observed for Arabic as task complexity increases.\\

\subsection{Reliability of Confidence \& Explanations}

The negative relationship between model-reported confidence and correctness suggests that self-assessed confidence reflects surface-level fluency rather than medical correctness. In multiple-choice settings, this can lead to confident selection of plausible but incorrect options, limiting the usefulness of confidence as an indicator of output reliability. While this effect appears across languages, it is particularly problematic in low-resource settings, where lower baseline accuracy increases the risk of over-trusting incorrect outputs.

Our explanation prompting results caution against treating generated rationales as a reliable remedy. Rather than improving outcomes, explanations induce a model- and language-dependent behavioral shift that reallocates generation toward coherent justifications instead of answer selection. These findings show that self-reported confidence and free-form explanations are insufficient as stand-alone reliability signals for multilingual medical QA, motivating evaluation and calibration approaches beyond model-internal self-assessments.

\section{Conclusion and Future Work}

Our findings underscore the need for language-aware adaptation across the entire modeling pipeline. At the representation level, tokenization must better capture morphological and domain-specific structure; at evaluation, alignment constraints should avoid conflating surface-form variation with reasoning errors; and at deployment, stronger calibration is required, as model confidence and explanations are unreliable in multilingual medical settings. Overall, these results suggest that improving medical LLM performance in underrepresented languages requires coordinated design choices rather than isolated model scaling or domain specialization.

More broadly, we present a diagnostic evaluation framework that combines controlled cross-lingual comparisons, question-level analysis, and reliability assessment to expose systematic weaknesses obscured by aggregate accuracy metrics. Although our study focuses on Arabic–English medical tasks, the methodology is language-agnostic and applicable to other low-resource or multilingual settings. We hope this work encourages evaluation protocols that explicitly account for linguistic structure, robustness, and reliability when developing medical AI systems for diverse clinical populations.

\section*{Acknlowedgements}
This work was supported by the Meem Foundation and the New York University Abu Dhabi (NYUAD) Center for Interdisciplinary Data Science and AI (CIDSAI), funded by Tamkeen under the NYUAD Research Institute Award CG016. The research was carried out on NYUAD's High Performance Computing resources (Jubail).

\section*{Limitations}
This study examines Arabic and English as a controlled language pair, with Arabic representing a widely spoken yet underrepresented language in medical NLP. While Arabic presents distinct linguistic and morphological challenges, it does not reflect the full diversity of low-resource or typologically distant languages; therefore, the generalizability of our findings beyond this pairing remains an open question.

Our analysis is diagnostic rather than causal. We identify systematic performance patterns across language, task complexity, tokenization behavior, and reliability signals, but do not isolate the effects of specific architectural choices, pre-training strategies, or data composition. This limitation is exacerbated by limited transparency in recent LLM training pipelines, which precludes controlled comparisons between adaptation paradigms such as instruction fine-tuning and continued pre-training.

Several design choices may also influence the results. Large models (70B) are evaluated using 4-bit quantization, which may introduce size-dependent effects but is required for evaluation at scale. In the explanation generation experiment, fixed generation budgets may disproportionately constrain Arabic outputs due to higher tokenization fragmentation; exploring language-adaptive generation limits is left to future work.

Finally, Arabic evaluations involve mixed-language prompts. While fully Arabic prompting was preliminarily tested and yielded lower performance, a systematic comparison of prompt-language strategies was out of scope. English versions of the dataset were obtained via automatic translation and were not manually validated; although the goal is cross-lingual comparison rather than translation quality assessment, translation noise may affect English performance.

\section*{Ethical Considerations}
This work evaluates LLMs for medical question answering, which carries inherent risks if such systems are deployed without appropriate safeguards. Our study is strictly evaluative and does not advocate the use of LLMs as standalone clinical decision-making tools. The dataset used in this study consists of non-patient-specific medical questions and does not involve real clinical records or personal health information. Furthermore, by highlighting systematic performance disparities across languages, this work aims to support more equitable evaluation and development of medical AI systems. 

\bibliography{custom}

@inproceedings{jeong2025,
    title = "Medical {A}daptation of {L}arge {L}anguage and {V}ision-{L}anguage {M}odels: {A}re {W}e {M}aking {P}rogress?",
    author = "Jeong, Daniel P  and
      Garg, Saurabh  and
      Lipton, Zachary Chase  and
      Oberst, Michael",
    editor = "Al-Onaizan, Yaser  and
      Bansal, Mohit  and
      Chen, Yun-Nung",
    booktitle = "Proceedings of the 2024 Conference on Empirical Methods in Natural Language Processing",
    month = nov,
    year = "2024",
    address = "Miami, Florida, USA",
    publisher = "Association for Computational Linguistics",
    url = "https://aclanthology.org/2024.emnlp-main.677/",
    doi = "10.18653/v1/2024.emnlp-main.677",
    pages = "12143--12170"
}

@inproceedings{wei2023,
author = {Wei, Jason and Wang, Xuezhi and Schuurmans, Dale and Bosma, Maarten and Ichter, Brian and Xia, Fei and Chi, Ed H. and Le, Quoc V. and Zhou, Denny},
title = {Chain-of-{T}hought {P}rompting {E}licits {R}easoning in {L}arge {L}anguage {M}odels},
year = {2022},
isbn = {9781713871088},
publisher = {Curran Associates Inc.},
address = {Red Hook, NY, USA},
booktitle = {Proceedings of the 36th International Conference on Neural Information Processing Systems},
articleno = {1800},
numpages = {14},
location = {New Orleans, LA, USA},
series = {NIPS '22}
}

@misc{nori2023,
      title={Capabilities of {GPT}-4 on {M}edical {C}hallenge {P}roblems}, 
      author={Harsha Nori and Nicholas King and Scott Mayer McKinney and Dean Carignan and Eric Horvitz},
      year={2023},
      eprint={2303.13375},
      archivePrefix={arXiv},
      primaryClass={cs.CL},
      url={https://arxiv.org/abs/2303.13375}, 
}

@article{singhal2023,
  title   = {Toward {E}xpert-{L}evel {M}edical {Q}uestion {A}nswering with {L}arge {L}anguage {M}odels},
  author  = {Singhal, Karan and Tu, Tao and Gottweis, Juraj and Sayres, Rory and
             Wulczyn, Ellery and Amin, Mohamed and Hou, Le and Clark, Kevin and
             Pfohl, Stephen R. and Cole-Lewis, Heather and Neal, Darlene and
             Rashid, Qazi Mamunur and Schaekermann, Mike and Wang, Amy and
             Dash, Dev and Chen, Jonathan H. and Shah, Nigam H. and
             Lachgar, Sami and Mansfield, Philip Andrew and Prakash, Sushant and
             Green, Bradley and Dominowska, Ewa and Ag{\"u}era y Arcas, Blaise and
             Toma{\v{s}}ev, Nenad and Liu, Yun and Wong, Renee and
             Semturs, Christopher and Mahdavi, S. Sara and Barral, Joelle K. and
             Webster, Dale R. and Corrado, Greg S. and Matias, Yossi and
             Azizi, Shekoofeh and Karthikesalingam, Alan and Natarajan, Vivek},
  journal = {Nature Medicine},
  volume  = {31},
  number  = {3},
  pages   = {943--950},
  year    = {2025},
  doi     = {10.1038/s41591-024-03423-7},
  url     = {https://doi.org/10.1038/s41591-024-03423-7}
}

@misc{wu2025,
      title={Med{R}eason: {E}liciting {F}actual {M}edical {R}easoning {S}teps in {LLM}s via {K}nowledge {G}raphs}, 
      author={Juncheng Wu and Wenlong Deng and Xingxuan Li and Sheng Liu and Taomian Mi and Yifan Peng and Ziyang Xu and Yi Liu and Hyunjin Cho and Chang-In Choi and Yihan Cao and Hui Ren and Xiang Li and Xiaoxiao Li and Yuyin Zhou},
      year={2025},
      eprint={2504.00993},
      archivePrefix={arXiv},
      primaryClass={cs.CL},
      url={https://arxiv.org/abs/2504.00993}, 
}

@misc{abudaoud2026medarabenchlargescalearabicmedical,
      title={MedAraBench: Large-Scale Arabic Medical Question Answering Dataset and Benchmark}, 
      author={Mouath Abu-Daoud and Leen Kharouf and Omar El Hajj and Dana El Samad and Mariam Al-Omari and Jihad Mallat and Khaled Saleh and Nizar Habash and Farah E. Shamout},
      year={2026},
      eprint={2602.01714},
      archivePrefix={arXiv},
      primaryClass={cs.CL},
      url={https://arxiv.org/abs/2602.01714}, 
}

@inproceedings{chen2025,
    title = "{T}owards {M}edical {C}omplex {R}easoning with {LLM}s through {M}edical {V}erifiable {P}roblems",
    author = "Chen, Junying  and
      Cai, Zhenyang  and
      Ji, Ke  and
      Wang, Xidong  and
      Liu, Wanlong  and
      Wang, Rongsheng  and
      Wang, Benyou",
    editor = "Che, Wanxiang  and
      Nabende, Joyce  and
      Shutova, Ekaterina  and
      Pilehvar, Mohammad Taher",
    booktitle = "Findings of the Association for Computational Linguistics: ACL 2025",
    month = jul,
    year = "2025",
    address = "Vienna, Austria",
    publisher = "Association for Computational Linguistics",
    url = "https://aclanthology.org/2025.findings-acl.751/",
    doi = "10.18653/v1/2025.findings-acl.751",
    pages = "14552--14573",
    ISBN = "979-8-89176-256-5"
}

@inproceedings{zhang2018,
author = {Zhang, Xiao and Wu, Ji and He, Zhiyang and Liu, Xien and Su, Ying},
title = {Medical {E}xam {Q}uestion {A}nswering with {L}arge-{S}cale {R}eading {C}omprehension},
year = {2018},
isbn = {978-1-57735-800-8},
publisher = {AAAI Press},
booktitle = {Proceedings of the Thirty-Second AAAI Conference on Artificial Intelligence and Thirtieth Innovative Applications of Artificial Intelligence Conference and Eighth AAAI Symposium on Educational Advances in Artificial Intelligence},
articleno = {700},
numpages = {8},
location = {New Orleans, Louisiana, USA},
series = {AAAI'18/IAAI'18/EAAI'18}
}

@InProceedings{pal2022,
  title = 	 {Med{MCQA}: {A} {L}arge-scale {M}ulti-{S}ubject {M}ulti-{C}hoice {D}ataset for {M}edical domain {Q}uestion {A}nswering},
  author =       {Pal, Ankit and Umapathi, Logesh Kumar and Sankarasubbu, Malaikannan},
  booktitle = 	 {Proceedings of the Conference on Health, Inference, and Learning},
  pages = 	 {248--260},
  year = 	 {2022},
  editor = 	 {Flores, Gerardo and Chen, George H and Pollard, Tom and Ho, Joyce C and Naumann, Tristan},
  volume = 	 {174},
  series = 	 {Proceedings of Machine Learning Research},
  month = 	 {07--08 Apr},
  publisher =    {PMLR},
  url = 	 {https://proceedings.mlr.press/v174/pal22a.html}
}

@inproceedings{singh2025,
    title = "Global {MMLU}: {U}nderstanding and {A}ddressing {C}ultural and {L}inguistic {B}iases in {M}ultilingual {E}valuation",
    author = "Singh, Shivalika  and
      Romanou, Angelika  and
      Fourrier, Cl{\'e}mentine  and
      Adelani, David Ifeoluwa  and
      Ngui, Jian Gang  and
      Vila-Suero, Daniel  and
      Limkonchotiwat, Peerat  and
      Marchisio, Kelly  and
      Leong, Wei Qi  and
      Susanto, Yosephine  and
      Ng, Raymond  and
      Longpre, Shayne  and
      Ruder, Sebastian  and
      Ko, Wei-Yin  and
      Bosselut, Antoine  and
      Oh, Alice  and
      Martins, Andre  and
      Choshen, Leshem  and
      Ippolito, Daphne  and
      Ferrante, Enzo  and
      Fadaee, Marzieh  and
      Ermis, Beyza  and
      Hooker, Sara",
    editor = "Che, Wanxiang  and
      Nabende, Joyce  and
      Shutova, Ekaterina  and
      Pilehvar, Mohammad Taher",
    booktitle = "Proceedings of the 63rd Annual Meeting of the Association for Computational Linguistics (Volume 1: Long Papers)",
    month = jul,
    year = "2025",
    address = "Vienna, Austria",
    publisher = "Association for Computational Linguistics",
    url = "https://aclanthology.org/2025.acl-long.919/",
    doi = "10.18653/v1/2025.acl-long.919",
    pages = "18761--18799",
    ISBN = "979-8-89176-251-0"
}

@inproceedings{jin2019,
    title = "{P}ub{M}ed{QA}: {A} {D}ataset for {B}iomedical {R}esearch {Q}uestion {A}nswering",
    author = "Jin, Qiao  and
      Dhingra, Bhuwan  and
      Liu, Zhengping  and
      Cohen, William  and
      Lu, Xinghua",
    editor = "Inui, Kentaro  and
      Jiang, Jing  and
      Ng, Vincent  and
      Wan, Xiaojun",
    booktitle = "Proceedings of the 2019 Conference on Empirical Methods in Natural Language Processing and the 9th International Joint Conference on Natural Language Processing (EMNLP-IJCNLP)",
    month = nov,
    year = "2019",
    address = "Hong Kong, China",
    publisher = "Association for Computational Linguistics",
    url = "https://aclanthology.org/D19-1259/",
    doi = "10.18653/v1/D19-1259",
    pages = "2567--2577"
}

@Article{jin2020,
AUTHOR = {Jin, Di and Pan, Eileen and Oufattole, Nassim and Weng, Wei-Hung and Fang, Hanyi and Szolovits, Peter},
TITLE = {What {D}isease {D}oes {T}his {P}atient {H}ave? {A} {L}arge-{S}ale {O}pen {D}omain {Q}uestion {A}nswering {D}ataset from {M}edical {E}xams},
JOURNAL = {Applied Sciences},
VOLUME = {11},
YEAR = {2021},
NUMBER = {14},
ARTICLE-NUMBER = {6421},
URL = {https://www.mdpi.com/2076-3417/11/14/6421},
ISSN = {2076-3417},
DOI = {10.3390/app11146421}
}

@misc{nori2023generalist,
      title={Can {G}eneralist {F}oundation {M}odels {O}utcompete {S}pecial-{P}urpose {T}uning? {C}ase {S}tudy in {M}edicine}, 
      author={Harsha Nori and Yin Tat Lee and Sheng Zhang and Dean Carignan and Richard Edgar and Nicolo Fusi and Nicholas King and Jonathan Larson and Yuanzhi Li and Weishung Liu and Renqian Luo and Scott Mayer McKinney and Robert Osazuwa Ness and Hoifung Poon and Tao Qin and Naoto Usuyama and Chris White and Eric Horvitz},
      year={2023},
      eprint={2311.16452},
      archivePrefix={arXiv},
      primaryClass={cs.CL},
      url={https://arxiv.org/abs/2311.16452}, 
}

@inproceedings{labrak2024-biomistral,
    title = "{B}io{M}istral: {A} {C}ollection of {O}pen-{S}ource {P}retrained {L}arge {L}anguage {M}odels for {M}edical {D}omains",
    author = "Labrak, Yanis  and
      Bazoge, Adrien  and
      Morin, Emmanuel  and
      Gourraud, Pierre-Antoine  and
      Rouvier, Mickael  and
      Dufour, Richard",
    editor = "Ku, Lun-Wei  and
      Martins, Andre  and
      Srikumar, Vivek",
    booktitle = "Findings of the Association for Computational Linguistics: ACL 2024",
    month = aug,
    year = "2024",
    address = "Bangkok, Thailand",
    publisher = "Association for Computational Linguistics",
    url = "https://aclanthology.org/2024.findings-acl.348/",
    doi = "10.18653/v1/2024.findings-acl.348",
    pages = "5848--5864"
}

@inproceedings{xiong2024hallucinate,
    title = "Benchmarking {R}etrieval-{A}ugmented {G}eneration for {M}edicine",
    author = "Xiong, Guangzhi  and
      Jin, Qiao  and
      Lu, Zhiyong  and
      Zhang, Aidong",
    editor = "Ku, Lun-Wei  and
      Martins, Andre  and
      Srikumar, Vivek",
    booktitle = "Findings of the Association for Computational Linguistics: ACL 2024",
    month = aug,
    year = "2024",
    address = "Bangkok, Thailand",
    publisher = "Association for Computational Linguistics",
    url = "https://aclanthology.org/2024.findings-acl.372/",
    doi = "10.18653/v1/2024.findings-acl.372",
    pages = "6233--6251"
}

@misc{med42,
      title={Med42-v2: A {S}uite of {C}linical {LLM}s}, 
      author={Clément Christophe and Praveen K Kanithi and Tathagata Raha and Shadab Khan and Marco AF Pimentel},
      year={2024},
      eprint={2408.06142},
      archivePrefix={arXiv},
      primaryClass={cs.CL},
      url={https://arxiv.org/abs/2408.06142}, 
}

@misc{deepseekv32,
      title={DeepSeek-V3.2: {P}ushing the {F}rontier of {O}pen {L}arge {L}anguage {M}odels}, 
      author={DeepSeek-AI and Aixin Liu and Aoxue Mei and Bangcai Lin and Bing Xue and Bingxuan Wang and Bingzheng Xu and Bochao Wu and Bowei Zhang and Chaofan Lin and Chen Dong and Chengda Lu and Chenggang Zhao and Chengqi Deng and Chenhao Xu and Chong Ruan and Damai Dai and Daya Guo and Dejian Yang and Deli Chen and Erhang Li and Fangqi Zhou and Fangyun Lin and Fucong Dai and Guangbo Hao and Guanting Chen and Guowei Li and H. Zhang and Hanwei Xu and Hao Li and Haofen Liang and Haoran Wei and Haowei Zhang and Haowen Luo and Haozhe Ji and Honghui Ding and Hongxuan Tang and Huanqi Cao and Huazuo Gao and Hui Qu and Hui Zeng and Jialiang Huang and Jiashi Li and Jiaxin Xu and Jiewen Hu and Jingchang Chen and Jingting Xiang and Jingyang Yuan and Jingyuan Cheng and Jinhua Zhu and Jun Ran and Junguang Jiang and Junjie Qiu and Junlong Li and Junxiao Song and Kai Dong and Kaige Gao and Kang Guan and Kexin Huang and Kexing Zhou and Kezhao Huang and Kuai Yu and Lean Wang and Lecong Zhang and Lei Wang and Liang Zhao and Liangsheng Yin and Lihua Guo and Lingxiao Luo and Linwang Ma and Litong Wang and Liyue Zhang and M. S. Di and M. Y Xu and Mingchuan Zhang and Minghua Zhang and Minghui Tang and Mingxu Zhou and Panpan Huang and Peixin Cong and Peiyi Wang and Qiancheng Wang and Qihao Zhu and Qingyang Li and Qinyu Chen and Qiushi Du and Ruiling Xu and Ruiqi Ge and Ruisong Zhang and Ruizhe Pan and Runji Wang and Runqiu Yin and Runxin Xu and Ruomeng Shen and Ruoyu Zhang and S. H. Liu and Shanghao Lu and Shangyan Zhou and Shanhuang Chen and Shaofei Cai and Shaoyuan Chen and Shengding Hu and Shengyu Liu and Shiqiang Hu and Shirong Ma and Shiyu Wang and Shuiping Yu and Shunfeng Zhou and Shuting Pan and Songyang Zhou and Tao Ni and Tao Yun and Tian Pei and Tian Ye and Tianyuan Yue and Wangding Zeng and Wen Liu and Wenfeng Liang and Wenjie Pang and Wenjing Luo and Wenjun Gao and Wentao Zhang and Xi Gao and Xiangwen Wang and Xiao Bi and Xiaodong Liu and Xiaohan Wang and Xiaokang Chen and Xiaokang Zhang and Xiaotao Nie and Xin Cheng and Xin Liu and Xin Xie and Xingchao Liu and Xingkai Yu and Xingyou Li and Xinyu Yang and Xinyuan Li and Xu Chen and Xuecheng Su and Xuehai Pan and Xuheng Lin and Xuwei Fu and Y. Q. Wang and Yang Zhang and Yanhong Xu and Yanru Ma and Yao Li and Yao Li and Yao Zhao and Yaofeng Sun and Yaohui Wang and Yi Qian and Yi Yu and Yichao Zhang and Yifan Ding and Yifan Shi and Yiliang Xiong and Ying He and Ying Zhou and Yinmin Zhong and Yishi Piao and Yisong Wang and Yixiao Chen and Yixuan Tan and Yixuan Wei and Yiyang Ma and Yiyuan Liu and Yonglun Yang and Yongqiang Guo and Yongtong Wu and Yu Wu and Yuan Cheng and Yuan Ou and Yuanfan Xu and Yuduan Wang and Yue Gong and Yuhan Wu and Yuheng Zou and Yukun Li and Yunfan Xiong and Yuxiang Luo and Yuxiang You and Yuxuan Liu and Yuyang Zhou and Z. F. Wu and Z. Z. Ren and Zehua Zhao and Zehui Ren and Zhangli Sha and Zhe Fu and Zhean Xu and Zhenda Xie and Zhengyan Zhang and Zhewen Hao and Zhibin Gou and Zhicheng Ma and Zhigang Yan and Zhihong Shao and Zhixian Huang and Zhiyu Wu and Zhuoshu Li and Zhuping Zhang and Zian Xu and Zihao Wang and Zihui Gu and Zijia Zhu and Zilin Li and Zipeng Zhang and Ziwei Xie and Ziyi Gao and Zizheng Pan and Zongqing Yao and Bei Feng and Hui Li and J. L. Cai and Jiaqi Ni and Lei Xu and Meng Li and Ning Tian and R. J. Chen and R. L. Jin and S. S. Li and Shuang Zhou and Tianyu Sun and X. Q. Li and Xiangyue Jin and Xiaojin Shen and Xiaosha Chen and Xinnan Song and Xinyi Zhou and Y. X. Zhu and Yanping Huang and Yaohui Li and Yi Zheng and Yuchen Zhu and Yunxian Ma and Zhen Huang and Zhipeng Xu and Zhongyu Zhang and Dongjie Ji and Jian Liang and Jianzhong Guo and Jin Chen and Leyi Xia and Miaojun Wang and Mingming Li and Peng Zhang and Ruyi Chen and Shangmian Sun and Shaoqing Wu and Shengfeng Ye and T. Wang and W. L. Xiao and Wei An and Xianzu Wang and Xiaowen Sun and Xiaoxiang Wang and Ying Tang and Yukun Zha and Zekai Zhang and Zhe Ju and Zhen Zhang and Zihua Qu},
      year={2025},
      eprint={2512.02556},
      archivePrefix={arXiv},
      primaryClass={cs.CL},
      url={https://arxiv.org/abs/2512.02556}, 
}

@inproceedings{vladika2025outdated,
    title = "Facts {F}ade {F}ast: {E}valuating {M}emorization of {O}utdated {M}edical {K}nowledge in {L}arge {L}anguage {M}odels",
    author = "Vladika, Juraj  and
      Dhaini, Mahdi  and
      Matthes, Florian",
    editor = "Christodoulopoulos, Christos  and
      Chakraborty, Tanmoy  and
      Rose, Carolyn  and
      Peng, Violet",
    booktitle = "Findings of the Association for Computational Linguistics: EMNLP 2025",
    month = nov,
    year = "2025",
    address = "Suzhou, China",
    publisher = "Association for Computational Linguistics",
    url = "https://aclanthology.org/2025.findings-emnlp.487/",
    doi = "10.18653/v1/2025.findings-emnlp.487",
    pages = "9161--9174",
    ISBN = "979-8-89176-335-7"
}

@article{qiu2024multi,
  title={Towards building multilingual language model for medicine},
  author={Qiu, Pengcheng and Wu, Chaoyi and Zhang, Xiaoman and Lin, Weixiong and Wang, Haicheng and Zhang, Ya and Wang, Yanfeng and Xie, Weidi},
  journal={Nature Communications},
  volume={15},
  number={1},
  pages={8384},
  year={2024},
  publisher={Nature Publishing Group UK London}
}

@misc{kembu2025italian,
      title={Are {LLM}s {T}ruly {M}ultilingual? {E}xploring {Z}ero-{S}hot {M}ultilingual {C}apability of {LLM}s for {I}nformation {R}etrieval: {A}n {I}talian {H}ealthcare {U}se {C}ase}, 
      author={Vignesh Kumar Kembu and Pierandrea Morandini and Marta Bianca Maria Ranzini and Antonino Nocera},
      year={2025},
      eprint={2512.04834},
      archivePrefix={arXiv},
      primaryClass={cs.AI},
      url={https://arxiv.org/abs/2512.04834}, 
}

@inproceedings{jin2023,
author = {Jin, Yiqiao and Chandra, Mohit and Verma, Gaurav and Hu, Yibo and De Choudhury, Munmun and Kumar, Srijan},
title = {Better to {A}sk in {E}nglish: {C}ross-{L}ingual {E}valuation of {L}arge {L}anguage {M}odels for {H}ealthcare {Q}ueries},
year = {2024},
isbn = {9798400701719},
publisher = {Association for Computing Machinery},
address = {New York, NY, USA},
url = {https://doi.org/10.1145/3589334.3645643},
doi = {10.1145/3589334.3645643},
booktitle = {Proceedings of the ACM Web Conference 2024},
pages = {2627–2638},
numpages = {12},
location = {Singapore, Singapore},
series = {WWW '24}
}

@misc{fanarteam2025,
      title={Fanar: An {A}rabic-{C}entric {M}ultimodal {G}enerative {AI} {P}latform}, 
      author={Fanar Team and Ummar Abbas and Mohammad Shahmeer Ahmad and Firoj Alam and Enes Altinisik and Ehsannedin Asgari and Yazan Boshmaf and Sabri Boughorbel and Sanjay Chawla and Shammur Chowdhury and Fahim Dalvi and Kareem Darwish and Nadir Durrani and Mohamed Elfeky and Ahmed Elmagarmid and Mohamed Eltabakh and Masoomali Fatehkia and Anastasios Fragkopoulos and Maram Hasanain and Majd Hawasly and Mus'ab Husaini and Soon-Gyo Jung and Ji Kim Lucas and Walid Magdy and Safa Messaoud and Abubakr Mohamed and Tasnim Mohiuddin and Basel Mousi and Hamdy Mubarak and Ahmad Musleh and Zan Naeem and Mourad Ouzzani and Dorde Popovic and Amin Sadeghi and Husrev Taha Sencar and Mohammed Shinoy and Omar Sinan and Yifan Zhang and Ahmed Ali and Yassine El Kheir and Xiaosong Ma and Chaoyi Ruan},
      year={2025},
      eprint={2501.13944},
      archivePrefix={arXiv},
      primaryClass={cs.CL},
      url={https://arxiv.org/abs/2501.13944}, 
}

@misc{mistral_small32_24b,
  title        = {Mistral-Small-3.2-24B-Instruct-2506},
  author       = {{Mistral AI}},
  year         = {2025},
  howpublished = {\url{https://huggingface.co/mistralai/Mistral-Small-3.2-24B-Instruct-2506}},
  note         = {Hugging Face model card.}
}

@misc{meditron3_70b_2024,
  title        = {Llama-3.1 Meditron-3 [70B]},
  author       = {{OpenMeditron Initiative}},
  year         = {2024},
  howpublished = {\url{https://huggingface.co/OpenMeditron/Meditron3-70B}},
  note         = {Hugging Face model card; publication forthcoming}
}

@misc{sequencematcher,
      title={When {A}re {N}ames {S}imilar {O}r the {S}ame? {I}ntroducing the {C}ode {N}ames {M}atcher {L}ibrary}, 
      author={Moshe Munk and Dror G. Feitelson},
      year={2022},
      eprint={2209.03198},
      archivePrefix={arXiv},
      primaryClass={cs.SE},
      url={https://arxiv.org/abs/2209.03198}, 
}

@misc{epochai,
  author       = {{Epoch AI}},
  title        = {Data on {N}otable {AI} {M}odels},
  year         = {2024},
  howpublished = {\url{https://epoch.ai/data/notable-ai-models}},
  note         = {Accessed: 2025-05-11}
}

@inproceedings{alasmari2024,
    title = "{A}ra{M}ed: {A}rabic {M}edical {Q}uestion {A}nswering using {P}retrained {T}ransformer {L}anguage {M}odels",
    author = "Alasmari, Ashwag  and
      Alhumoud, Sarah  and
      Alshammari, Waad",
    editor = "Al-Khalifa, Hend  and
      Darwish, Kareem  and
      Mubarak, Hamdy  and
      Ali, Mona  and
      Elsayed, Tamer",
    booktitle = "Proceedings of the 6th Workshop on Open-Source Arabic Corpora and Processing Tools (OSACT) with Shared Tasks on Arabic LLMs Hallucination and Dialect to MSA Machine Translation @ LREC-COLING 2024",
    month = may,
    year = "2024",
    address = "Torino, Italia",
    publisher = "ELRA and ICCL",
    url = "https://aclanthology.org/2024.osact-1.6/",
    pages = "50--56",
}

@misc{sellergren2025,
      title={Med{G}emma {T}echnical {R}eport}, 
      author={Andrew Sellergren and Sahar Kazemzadeh and Tiam Jaroensri and Atilla Kiraly and Madeleine Traverse and Timo Kohlberger and Shawn Xu and Fayaz Jamil and Cían Hughes and Charles Lau and Justin Chen and Fereshteh Mahvar and Liron Yatziv and Tiffany Chen and Bram Sterling and Stefanie Anna Baby and Susanna Maria Baby and Jeremy Lai and Samuel Schmidgall and Lu Yang and Kejia Chen and Per Bjornsson and Shashir Reddy and Ryan Brush and Kenneth Philbrick and Mercy Asiedu and Ines Mezerreg and Howard Hu and Howard Yang and Richa Tiwari and Sunny Jansen and Preeti Singh and Yun Liu and Shekoofeh Azizi and Aishwarya Kamath and Johan Ferret and Shreya Pathak and Nino Vieillard and Ramona Merhej and Sarah Perrin and Tatiana Matejovicova and Alexandre Ramé and Morgane Riviere and Louis Rouillard and Thomas Mesnard and Geoffrey Cideron and Jean-bastien Grill and Sabela Ramos and Edouard Yvinec and Michelle Casbon and Elena Buchatskaya and Jean-Baptiste Alayrac and Dmitry Lepikhin and Vlad Feinberg and Sebastian Borgeaud and Alek Andreev and Cassidy Hardin and Robert Dadashi and Léonard Hussenot and Armand Joulin and Olivier Bachem and Yossi Matias and Katherine Chou and Avinatan Hassidim and Kavi Goel and Clement Farabet and Joelle Barral and Tris Warkentin and Jonathon Shlens and David Fleet and Victor Cotruta and Omar Sanseviero and Gus Martins and Phoebe Kirk and Anand Rao and Shravya Shetty and David F. Steiner and Can Kirmizibayrak and Rory Pilgrim and Daniel Golden and Lin Yang},
      year={2025},
      eprint={2507.05201},
      archivePrefix={arXiv},
      primaryClass={cs.AI},
      url={https://arxiv.org/abs/2507.05201}, 
}

@misc{llama33_70b,
  title        = {LLaMA 3.3 70B Instruct},
  author       = {{Meta AI}},
  year         = {2024},
  howpublished = {\url{https://huggingface.co/meta-llama/Llama-3.3-70B-Instruct}},
  note         = {Hugging Face model card.}
}

@inproceedings{
bari2024,
title={{ALL}a{M}: {L}arge {L}anguage {M}odels for {A}rabic and {E}nglish},
author={M Saiful Bari and Yazeed Alnumay and Norah A. Alzahrani and Nouf M. Alotaibi and Hisham Abdullah Alyahya and Sultan AlRashed and Faisal Abdulrahman Mirza and Shaykhah Z. Alsubaie and Hassan A. Alahmed and Ghadah Alabduljabbar and Raghad Alkhathran and Yousef Almushayqih and Raneem Alnajim and Salman Alsubaihi and Maryam Al Mansour and Saad Amin Hassan and Dr. Majed Alrubaian and Ali Alammari and Zaki Alawami and Abdulmohsen Al-Thubaity and Ahmed Abdelali and Jeril Kuriakose and Abdalghani Abujabal and Nora Al-Twairesh and Areeb Alowisheq and Haidar Khan},
booktitle={The Thirteenth International Conference on Learning Representations},
year={2025},
url={https://openreview.net/forum?id=MscdsFVZrN}
}

@article{farghaly2009,
  title={Arabic natural language processing: Challenges and solutions},
  author={Farghaly, Ali and Shaalan, Khaled},
  journal={ACM Transactions on Asian Language Information Processing (TALIP)},
  volume={8},
  number={4},
  pages={1--22},
  year={2009},
  publisher={ACM New York, NY, USA}
}

@article{Alonso_2024,
   title={{M}ed{E}xp{QA}: Multilingual benchmarking of {L}arge {L}anguage {M}odels for {M}edical {Q}uestion {A}nswering},
   volume={155},
   ISSN={0933-3657},
   url={http://dx.doi.org/10.1016/j.artmed.2024.102938},
   DOI={10.1016/j.artmed.2024.102938},
   journal={Artificial Intelligence in Medicine},
   publisher={Elsevier BV},
   author={Alonso, Iñigo and Oronoz, Maite and Agerri, Rodrigo},
   year={2024},
   month=sep, pages={102938} }

@inproceedings{zhang2023,
    title = "{H}uatuo{GPT}, {T}owards {T}aming {L}anguage {M}odel to {B}e a {D}octor",
    author = "Zhang, Hongbo  and
      Chen, Junying  and
      Jiang, Feng  and
      Yu, Fei  and
      Chen, Zhihong  and
      Chen, Guiming  and
      Li, Jianquan  and
      Wu, Xiangbo  and
      Zhiyi, Zhang  and
      Xiao, Qingying  and
      Wan, Xiang  and
      Wang, Benyou  and
      Li, Haizhou",
    editor = "Bouamor, Houda  and
      Pino, Juan  and
      Bali, Kalika",
    booktitle = "Findings of the Association for Computational Linguistics: EMNLP 2023",
    month = dec,
    year = "2023",
    address = "Singapore",
    publisher = "Association for Computational Linguistics",
    url = "https://aclanthology.org/2023.findings-emnlp.725/",
    doi = "10.18653/v1/2023.findings-emnlp.725",
    pages = "10859--10885"
}

@misc{sengupta2023jais,
      title={Jais and {J}ais-chat: {A}rabic-{C}entric {F}oundation and {I}nstruction-{T}uned {O}pen {G}enerative {L}arge {L}anguage {M}odels}, 
      author={Neha Sengupta and Sunil Kumar Sahu and Bokang Jia and Satheesh Katipomu and Haonan Li and Fajri Koto and William Marshall and Gurpreet Gosal and Cynthia Liu and Zhiming Chen and Osama Mohammed Afzal and Samta Kamboj and Onkar Pandit and Rahul Pal and Lalit Pradhan and Zain Muhammad Mujahid and Massa Baali and Xudong Han and Sondos Mahmoud Bsharat and Alham Fikri Aji and Zhiqiang Shen and Zhengzhong Liu and Natalia Vassilieva and Joel Hestness and Andy Hock and Andrew Feldman and Jonathan Lee and Andrew Jackson and Hector Xuguang Ren and Preslav Nakov and Timothy Baldwin and Eric Xing},
      year={2023},
      eprint={2308.16149},
      archivePrefix={arXiv},
      primaryClass={cs.CL},
      url={https://arxiv.org/abs/2308.16149}, 
}

@misc{medarabiq,
      title={MedArabiQ: Benchmarking Large Language Models on Arabic Medical Tasks}, 
      author={Mouath Abu Daoud and Chaimae Abouzahir and Leen Kharouf and Walid Al-Eisawi and Nizar Habash and Farah E. Shamout},
      year={2025},
      eprint={2505.03427},
      archivePrefix={arXiv},
      primaryClass={cs.CL},
      url={https://arxiv.org/abs/2505.03427}, 
}

@book{Habash:2010:introduction,
	Author = {Habash, Nizar Y},
	Journal = {Synthesis Lectures on Human Language Technologies},
	Pages = {1--187},
	Publisher = {Morgan \& Claypool Publishers},
	Title = {Introduction to {A}rabic natural language processing},
	Volume = {3},
	Year = {2010}}

@inproceedings{Obeid:2020:camel,
    title = "{CAM}e{L} {T}ools: {A}n {O}pen {S}ource {P}ython {T}oolkit for {A}rabic {N}atural {L}anguage {P}rocessing",
    author = "Obeid, Ossama  and
      Zalmout, Nasser  and
      Khalifa, Salam  and
      Taji, Dima  and
      Oudah, Mai  and
      Alhafni, Bashar  and
      Inoue, Go  and
      Eryani, Fadhl  and
      Erdmann, Alexander  and
      Habash, Nizar",
    editor = "Calzolari, Nicoletta  and
      B{\'e}chet, Fr{\'e}d{\'e}ric  and
      Blache, Philippe  and
      Choukri, Khalid  and
      Cieri, Christopher  and
      Declerck, Thierry  and
      Goggi, Sara  and
      Isahara, Hitoshi  and
      Maegaard, Bente  and
      Mariani, Joseph  and
      Mazo, H{\'e}l{\`e}ne  and
      Moreno, Asuncion  and
      Odijk, Jan  and
      Piperidis, Stelios",
    booktitle = "Proceedings of the Twelfth Language Resources and Evaluation Conference",
    month = may,
    year = "2020",
    address = "Marseille, France",
    publisher = "European Language Resources Association",
    url = "https://aclanthology.org/2020.lrec-1.868",
    pages = "7022--7032",
    language = "English",
    ISBN = "979-10-95546-34-4",
}

@inproceedings{inoue-etal-2021-interplay,
    title = "The Interplay of Variant, Size, and Task Type in {A}rabic Pre-trained Language Models",
    author = "Inoue, Go  and
      Alhafni, Bashar  and
      Baimukan, Nurpeiis  and
      Bouamor, Houda  and
      Habash, Nizar",
    editor = "Habash, Nizar  and
      Bouamor, Houda  and
      Hajj, Hazem  and
      Magdy, Walid  and
      Zaghouani, Wajdi  and
      Bougares, Fethi  and
      Tomeh, Nadi  and
      Abu Farha, Ibrahim  and
      Touileb, Samia",
    booktitle = "Proceedings of the Sixth Arabic Natural Language Processing Workshop",
    month = apr,
    year = "2021",
    address = "Kyiv, Ukraine (Virtual)",
    publisher = "Association for Computational Linguistics",
    url = "https://aclanthology.org/2021.wanlp-1.10/",
    pages = "92--104"
}

\clearpage

\newpage
\appendix

\renewcommand{\thesection}{\Alph{section}}

\setcounter{table}{0}
\renewcommand{\thetable}{\thesection\arabic{table}}
\setcounter{figure}{0}
\renewcommand{\thefigure}{\thesection\arabic{figure}}

\section{Data Samples}
\begin{figure}[H]
    \centering
    \includegraphics[width=0.99\linewidth]{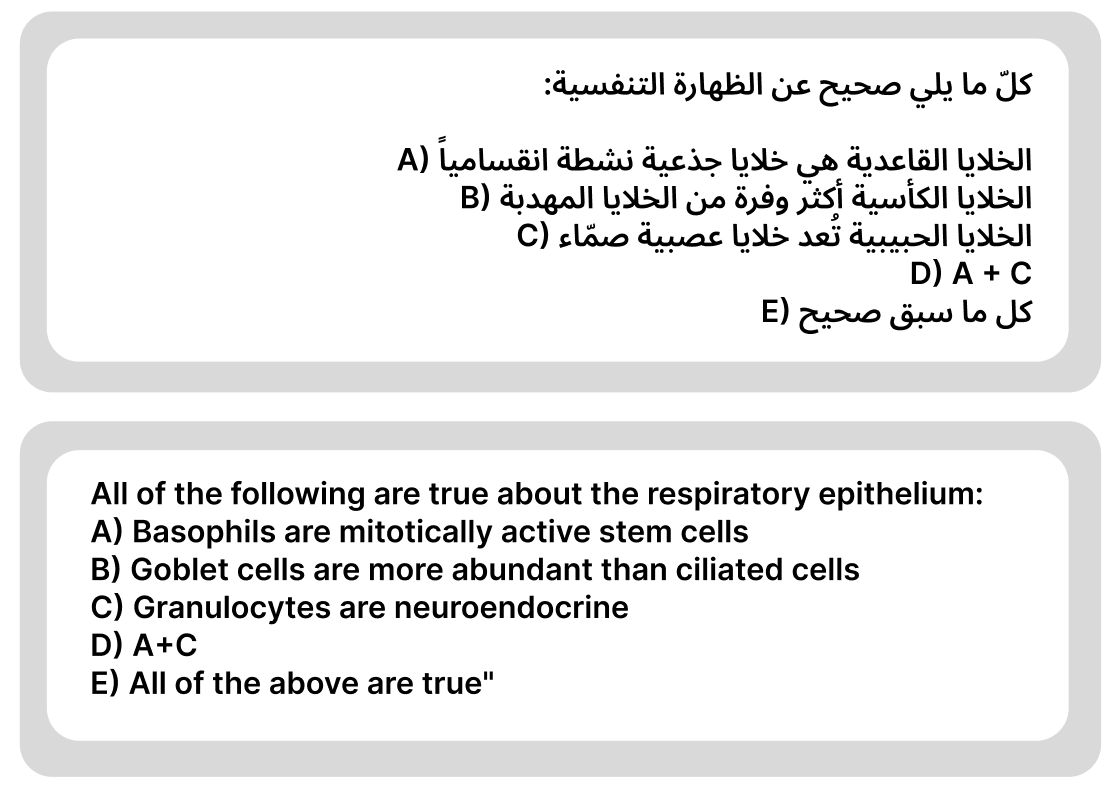}
    \caption{\textbf{Example dataset entry showing an Arabic MCQ (top) and its English translation (bottom).}}
    \label{fig:data-sample}
\end{figure}

Figure~\ref{fig:data-sample} shows representative examples from the dataset,
including the original Arabic question and its corresponding English translation,
to illustrate the structure and content of the bilingual data used in our experiments.

\section{Prompts Used}
\label{app:prompts}

\subsection{Letter-Based Prompting}
\label{app:prompt_letter}

\begin{figure}[H]
    \centering
    \includegraphics[width=0.99\linewidth]{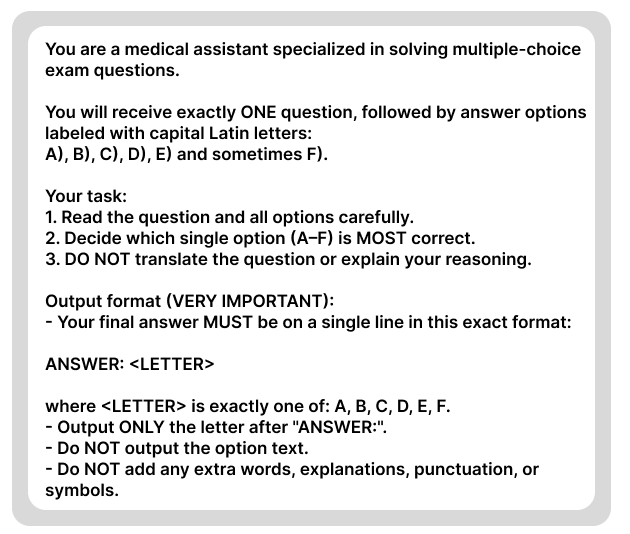}
    \caption{\textbf{Prompt template used for letter-based MCQ answering.}}
    \label{fig:prompt-letter}
\end{figure}

Figure~\ref{fig:prompt-letter} shows the exact prompt template used for
letter-based multiple-choice question answering, where the model is instructed
to return only a single answer option (A--F) without additional explanation.

\subsection{Text Generation Prompting }
\begin{figure}[H]
    \centering
    \includegraphics[width=0.99\linewidth]{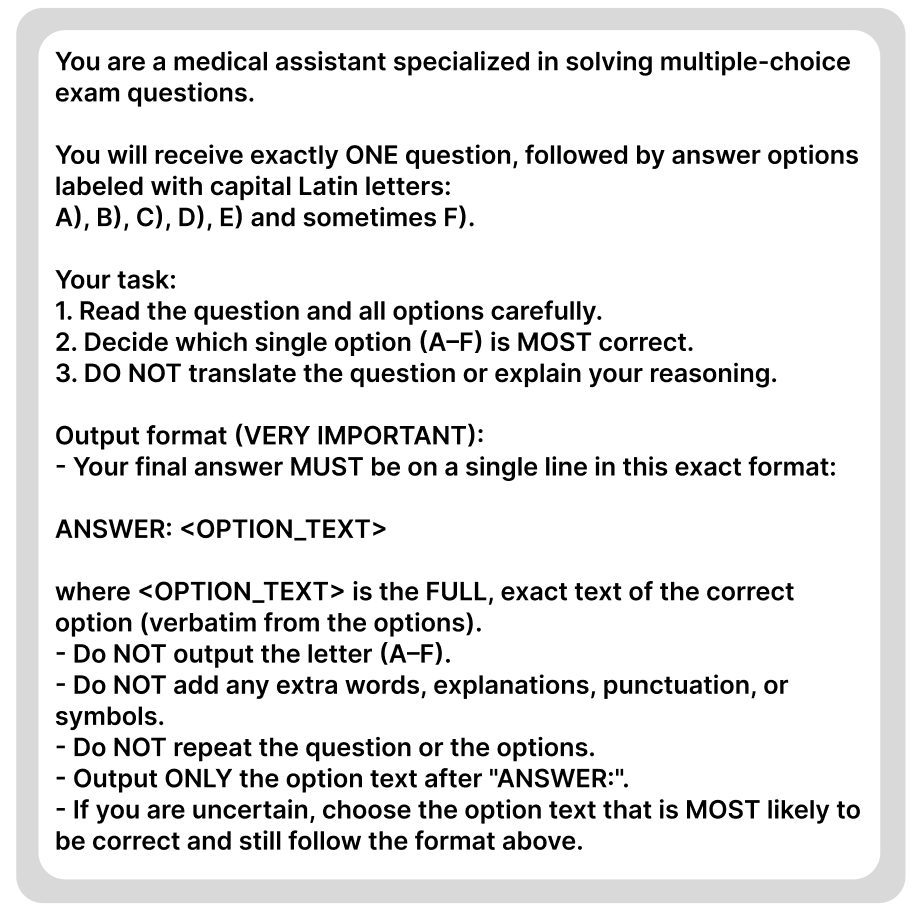}
    \caption{\textbf{Prompt template used for text generation MCQ answering.}}
    \label{fig:prompt-text}
\end{figure}

Figure~\ref{fig:prompt-text} shows the prompt template used for
exact text generation matching, where the model is instructed
to return the exact text sequence corresponding to the correct answer option without additional explanation.

\subsection{Confidence Generation Prompting }

\begin{figure}[H]
    \centering
    \includegraphics[width=0.99\linewidth]{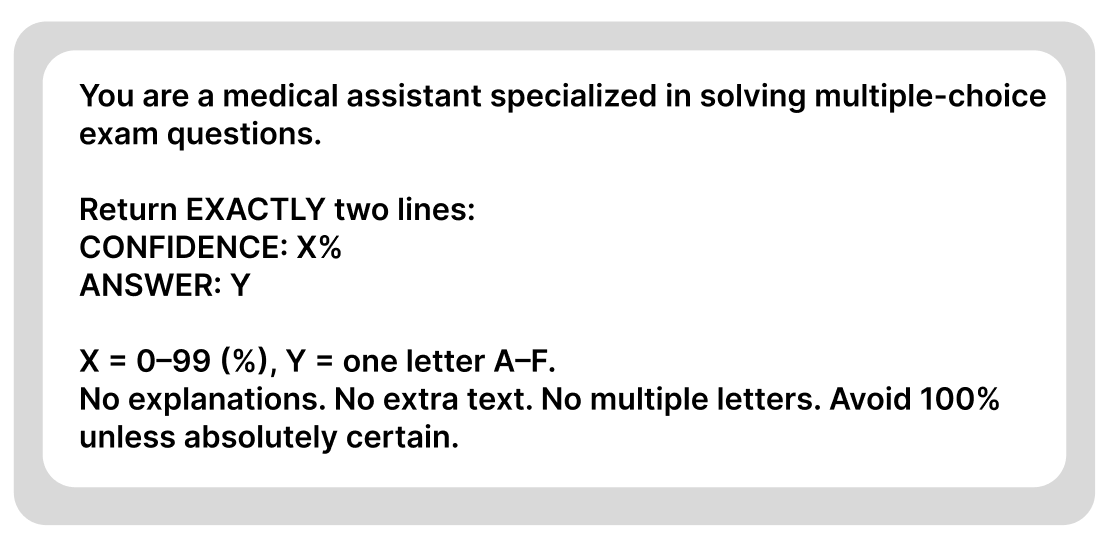}
    \caption{\textbf{Prompt template used for confidence generation MCQ answering.}}
    \label{fig:prompt-conf}
\end{figure}

Figure~\ref{fig:prompt-conf} shows the exact prompt template used for
confidence-aware MCQ answering, where the model is instructed
to report an explicit confidence estimate alongside its selected answer
using a strictly defined output format, without providing any additional
explanation.

\subsection{Explanation Generation Prompting}
\begin{figure}[H]
    \centering
    \includegraphics[width=0.99\linewidth]{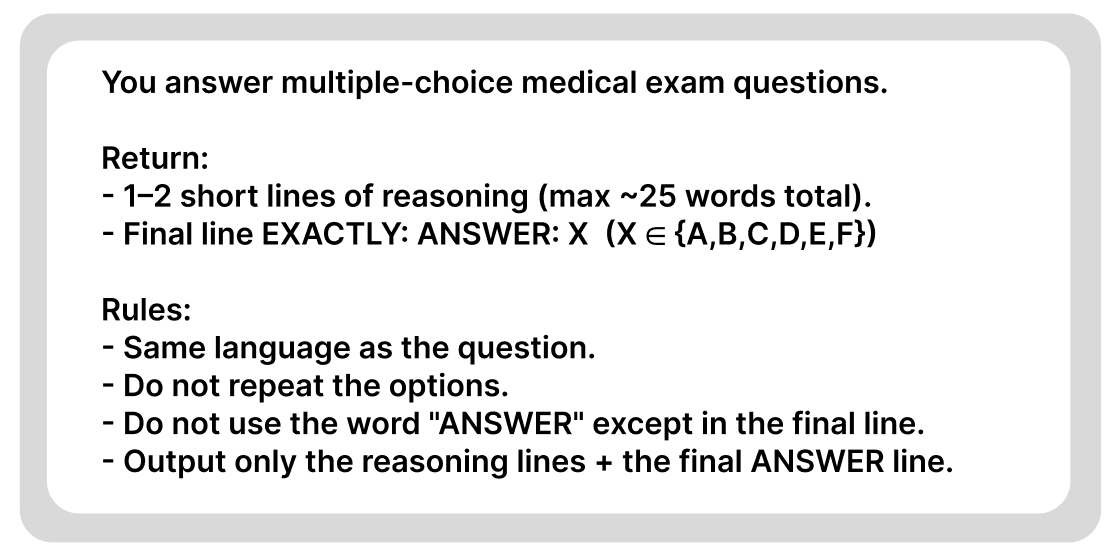}
    \caption{\textbf{Prompt template used for explanation-based MCQ answering.}}
    \label{fig:prompt-explanation}
\end{figure}

Figure~\ref{fig:prompt-explanation} shows the exact prompt template used for explanation-based MCQ answering, where the model is instructed
to generate a brief medical rationale followed by a final answer
selection, enabling analysis of whether generated explanations
correlate with answer correctness.

\section{Examples of Failure Modes Under Explanation Prompting}

\begin{figure}[t]
    \centering
    \includegraphics[width=0.99\linewidth]{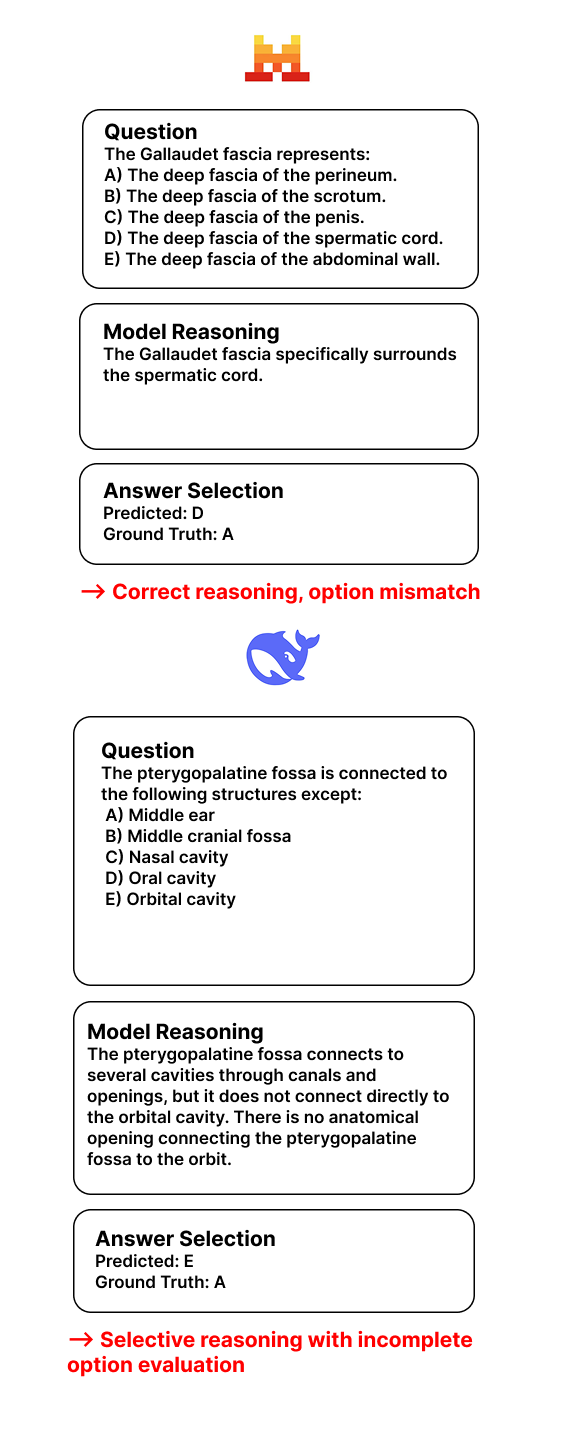}
    \caption{\textbf{Examples of reasoning–label misalignment under explanation prompting.} Models may produce correct or salient medical reasoning while selecting an incorrect option due to option mismatch or incomplete evaluation of alternatives.}
    \label{fig:expl-samples}
\end{figure}

Figure \ref{fig:expl-samples} shows representative examples of failure modes observed under explanation-conditioned prompting. In these cases, models generate medically plausible or partially correct explanations but select an incorrect answer option, revealing misalignment between reasoning and final answer selection.

\end{document}